# ACHIEVING GREATER EXPLANATORY POWER AND FORECASTING ACCURACY WITH NON-UNIFORM SPREAD FUZZY LINEAR REGRESSION

Arindam Chaudhuri[1,*] and Kajal De[2]

[1]Lecturer (Mathematics & Computer Science), Meghnad Saha Institute of Technology, Kolkata, India
[1]Research Scholar (Computer Science), Netaji Subhas Open University, Kolkata, India
arindam_chau@yahoo.co.in (*corresponding author)
[2]Professor in Mathematics, School of Science, Netaji Subhas Open University, Kolkata, India

**Abstract:** Fuzzy regression models have been applied to several Operations Research applications viz., forecasting and prediction. Earlier works on fuzzy regression analysis obtain crisp regression coefficients for eliminating the problem of increasing spreads for the estimated fuzzy responses as the magnitude of the independent variable increases. But they cannot deal with the problem of non-uniform spreads. In this work, a three-phase approach is discussed to construct the fuzzy regression model with non-uniform spreads to deal with this problem. The first phase constructs the membership functions of the least-squares estimates of regression coefficients based on extension principle to completely conserve the fuzziness of observations. They are then defuzzified by the centre of area method to obtain crisp regression coefficients in the second phase. Finally, the error terms of the method are determined by setting each estimated spread equal to its corresponding observed spread. The Tagaki-Sugeno inference system is used for improving the accuracy of forecasts. The simulation example demonstrates the strength of fuzzy linear regression model in terms of higher explanatory power and forecasting performance.

## 1. Introduction

Fuzzy regression is an important technique for analyzing the vague relationship between dependent variables (response variables) and independent variables (explanatory variables) in complex systems involving human subjective judgment under incomplete and imprecise data conditions, according to Kacprzyk and Fedrizzi (1992). Some successful applications of fuzzy regression include insurance, housing, thermal comfort forecasting, productivity and consumer satisfaction, product life cycle prediction, project evaluation, reservoir operations, actuarial analysis, robotic welding process and business cycle analysis. Tanaka et al. (1982) first studied fuzzy linear regression (FLR) problem with crisp explanatory variables and fuzzy response variables. They formulated FLR problem as linear programming model to determine regression coefficients as fuzzy numbers, where objective function minimizes total spread of fuzzy regression coefficients subject to constraint that the support of estimated values is needed to cover support of their associated observed values for certain pre-specified level. Later it was improved by Tanaka, Hayashi and Watada (1989). The drawbacks of these approaches have been pointed out by several investigations. Redden and Woodall (1994) stated the sensitiveness to outliers. Wang and Tsaur (2000) pointed wide ranges in estimation. Kao and Lin (2005) stated that more observations result in fuzzier estimations, which contradicts general observation that more observations provide better estimations. Some other works include fuzzy least-squares approach to determine regression coefficients as proposed by Diamond (1988) and criterion of minimizing difference of membership values between observed and estimated fuzzy dependent variable by Kim and Bishu (1998). Three types of multi-objective programming approaches to investigate FLR model with fuzzy explanatory variables and responses were formulated by Sakawa and Yano (1992). Hong et al. (2001) used shape preserving arithmetic operations LR fuzzy numbers for least-squares fitting to investigate class of FLR problem. Here, the derived regression coefficients are fuzzy numbers. However, since the regression coefficients derived based on Zadeh's extension principle are fuzzy numbers, spread of estimated dependent variable becomes wider as magnitudes of independent variables increase, even if spreads of observed dependent variables are actually decreasing. To avoid problem of wide spreads for large value of explanatory variables in estimation, Kao and Chyu (2002) proposed two-stage approach to obtain crisp regression coefficients in first stage and determine unique fuzzy error term in second stage. Kao and Chyu (2003) also proposed least-squares method to derive regression coefficients that are crisp. These two studies have better performance but they still cannot cope with the situation of decreasing or non-uniform spread. Another issue is that crisp regression coefficients may eliminate problem of increasing spread, but they also mislead functional relationship between dependent and independent variables in fuzzy environment. When spreads of fuzzy independent variables are large, it is possible that spread of regression coefficients is also large. In this case values of regression coefficients are in wide range even from negative to positive values. If derived regression coefficients are crisp, some valuable information may be lost. According to Bargiela et al. (2007) regression model based on fuzzy data shows beneficial characteristic of enhanced generalization of data patterns compared to regression models based on numeric data only. This is because membership function associated with fuzzy sets has significant informative value in terms of capturing either notion of accuracy of information or notion of proximity of patterns in data set used for derivation of regression model. Thus, when explanatory variables are fuzzy, regression coefficients will be fuzzy and they should be described by membership functions to completely conserve fuzziness of explanatory variables.

This work addresses two important problems of fuzzy linear regression viz., wide spreads for large value of explanatory variables and functional relationship between dependent and independent variables. First a procedure is developed for constructing membership function of fuzzy regression coefficients such that fuzziness of input information can be completely conserved. Then variable spread FLR model is generated with higher explanatory power and forecasting accuracy, which resolves problem of wide spreads of estimated response for larger values of independent variables in fuzzy regression such that the situation of decreasing or non-uniform spreads can be tackled. The above problems are taken care of by means of three-step approach. In first step, membership functions of least-squares estimation of fuzzy response and explanatory variables are derived based on Zadeh's extension principle given by Yager (1978), (1986) to completely conserve fuzziness such that some valuable information is obtained. In second step, fuzzy regression coefficients are defuzzified to crisp values via fuzzy ranking method to avoid problem of non-uniform spreads for larger values of explanatory variables in estimation. Finally, in third step, mathematical programming approach determines fuzzy error term for each pair of explanation variables and response, such that errors in estimation are minimized subject to constraints including spreads of each estimated response equal to that of associated response. As spreads of error terms coincide with those of their associated observed responses, spreads used here are non-uniform in nature, no matter how spreads of observed responses change. This paper is organized as follows. In the next section, FLR model is given. This is followed by a brief discussion on membership function on regression coefficients. In section 4, the concept of fuzzy regression model with non-uniform spreads is analyzed. A numerical example to illustrate the method is provided in section 5. Finally, in section 6 conclusions are given.

## 2. Fuzzy Linear Regression

The crisp simple linear regression model involving a single independent variable and one or more explanatory variable is expressed mathematically as $y_i = \beta_0 + \beta_1 x_i + \varepsilon_i, i = 1,\ldots,n$ (1) where $x_i$ and $y_i$ represent explanatory variable and the response in $i^{th}$ observation respectively; $\beta_0$ and $\beta_1$ are regression coefficients, and $\varepsilon_i$ is error term associated with $i^{th}$ observation. In classical statistical analysis, population parameters $\beta_0$ and $\beta_1$ are estimated by certain sample statistics, such as least-squares estimators, which are the best linear unbiased estimators used most frequently given by Rohatgi (1976) as:

$$b_1 = \frac{\sum_{i=1}^{n}(x_i - \bar{x})(y_i - \bar{y})}{\sum_{i=1}^{n}(x_i - \bar{x})^2} \quad (2i) \qquad b_0 = \frac{\sum_{i=1}^{n} y_i \sum_{i=1}^{n} x_i^2 - \sum_{i=1}^{n} x_i \sum_{i=1}^{n} x_i y_i}{n\sum_{i=1}^{n} x_i^2 - (\sum_{i=1}^{n} x_i)^2} = \bar{y} - b_1 \bar{x}, (2ii), \quad \text{where} \quad \bar{x} = \frac{\sum_{i=1}^{n} x_i}{n}$$

and $\bar{y} = \frac{\sum_{i=1}^{n} y_i}{n}$. The case of multiple regressions is straightforward generalization of the linear regression. When any of the responses $y_i$ or explanatory variables $x_i$ is fuzzy, crisp regression model defined in Equation (1) is then modified into fuzzy regression model expressed mathematically as $\tilde{y}_i = \tilde{\beta}_0 + \tilde{\beta}_1 \tilde{x}_i + \tilde{\varepsilon}_i, i = 1,\ldots,n$ (3) where, $\tilde{\beta}_0$ and $\tilde{\beta}_1$ are regression coefficients, $(\tilde{x}_i, \tilde{y}_i), i = 1,\ldots,n$ are $n$ pairs of fuzzy observations and $\tilde{\varepsilon}_i$ is fuzzy error term associated with the regression model. Since crisp numbers can be described by degenerated fuzzy numbers, consider $\tilde{x}_i$ and $\tilde{y}_i$ as fuzzy numbers as $\tilde{x}_i = \{(x_i, \mu_{\tilde{x}_i}(x_i)) \mid x_i \in X_i\}, i = 1,\ldots,n$ (4i) and

$\tilde{y}_i = \{(y_i, \mu_{\tilde{y}_i}(y_i)) \mid y_i \in Y_i\}, i = 1, \ldots, n$ (4ii) where $X_i$ and $Y_i$ are the crisp universal sets of the explanatory variables and responses, and $\mu_{\tilde{x}_i}(x_i)$ and $\mu_{\tilde{y}_i}(y_i)$ are their corresponding membership functions respectively.

When $\tilde{x}_i$ and $\tilde{y}_i$ are fuzzy numbers, the least-squares estimators stated in Equations (2i) and (2ii) become fuzzy numbers denoted as $\tilde{b}_1$ and $\tilde{b}_0$ respectively. It is necessary to find the membership functions to completely conserve the fuzziness of observations.

## 3. Membership Function of Regression Coefficients

According to Zadeh's Extension Principle, Yager (1978) and Zadeh (1986), the membership functions of regression coefficients are defined as $\mu_{\tilde{b}_1}(z_1) = \sup_{X_i, Y_i} \min\{\mu_{\tilde{x}_i}(x_i), \mu_{\tilde{y}_i}(y_i), \forall i \mid z_1 = b_1\}$ (5i) and $\mu_{\tilde{b}_0}(z_0) = \sup_{X_i, Y_i} \min\{\mu_{\tilde{x}_i}(x_i), \mu_{\tilde{y}_i}(y_i), \forall i \mid z_0 = b_0\}$ (5ii) where, $b_1$ and $b_0$ are defined in Equations (2i) and (2ii). Although Equations (5i) and (5ii) are theoretically correct, since several fuzzy numbers are involved, it is almost impossible to derive the membership functions of $\tilde{b}_1$ and $\tilde{b}_0$ from these two Equations in an explicit manner. An easier approach has been proposed by Chen (2004), (2005) to derive the membership function based on Zadeh's Extension Principle and $\alpha$ representation, which can be adopted here to construct the membership functions $\mu_{\tilde{b}_1}$ and $\mu_{\tilde{b}_0}$. The idea is to derive $\alpha$-cuts of $\tilde{b}_1$ and $\tilde{b}_0$ respectively which are defined for $\tilde{x}_i$ and $\tilde{y}_i$ given by Zimmermann (2001) and Klir et al. (2003) as: $x_i(\alpha) = \{x_i \in X_i \mid \mu_{\tilde{x}_i}(x_i) \geq \alpha\}$ (6i) and $y_i(\alpha) = \{y_i \in Y_i \mid \mu_{\tilde{y}_i}(y_i) \geq \alpha\}$ (6ii). Here, $x_i(\alpha)$ and $y_i(\alpha)$ are crisp sets rather than fuzzy sets. Since, $x_i$ and $y_i$ are assumed to be fuzzy numbers, their $\alpha$-cuts defined in Equations (5i) and (5ii) are crisp intervals expressed as follows: $x_i(\alpha) = [\min_{x_i \in X_i}\{x_i \mid \mu_{\tilde{x}_i}(x_i) \geq \alpha\}, \max_{x_i \in X_i}\{x_i \mid \mu_{\tilde{x}_i}(x_i) \geq \alpha\}] = [(x_i)_\alpha^L, (x_i)_\alpha^U]$ and $y_i(\alpha) = [\min_{y_i \in Y_i}\{y_i \mid \mu_{\tilde{y}_i}(y_i) \geq \alpha\}, \max_{y_i \in Y_i}\{y_i \mid \mu_{\tilde{y}_i}(y_i) \geq \alpha\}] = [(y_i)_\alpha^L, (y_i)_\alpha^U]$. These intervals indicate where the constant explanatory variables and responses lie at possibility $\alpha$. By convexity of a fuzzy number, the bounds of these intervals are functions of $\alpha$ and can be obtained $(x_i)_\alpha^L = \min \mu_{\tilde{x}_i}^{-1}(\alpha)$, $(x_i)_\alpha^U = \max \mu_{\tilde{x}_i}^{-1}(\alpha)$, $(y_i)_\alpha^L = \min \mu_{\tilde{y}_i}^{-1}(\alpha)$ and $(y_i)_\alpha^U = \max \mu_{\tilde{y}_i}^{-1}(\alpha)$ respectively. The membership functions of $b_1$ and $b_0$ defined in (5i) and (5ii) are parameterized by $\alpha$. As a result of this the $\alpha$-cuts can be used to construct their membership function.

The basic idea of deriving the membership function of the regression coefficients is to employ the $\alpha$-cuts and Zadeh's Extension Principle to transform the fuzzy regression coefficients $\tilde{b}_1$ and $\tilde{b}_0$ to a family of crisp regression coefficients $(\tilde{b}_1)_\alpha^L$ and $(\tilde{b}_0)_\alpha^L$, respectively parameterized by $\alpha$. According to the Extension Principle given by Equation (5i), $\mu_{\tilde{b}_1}(z_1)$ is minimum of $\mu_{\tilde{x}_i}(x_i)$ and $\mu_{\tilde{y}_i}(y_i), \forall i$. From the membership value, either $\mu_{\tilde{x}_i}(x_i) \geq \alpha$ or $\mu_{\tilde{y}_i}(y_i) \geq \alpha$, and at least one $\mu_{\tilde{x}_i}(x_i)$ or $\mu_{\tilde{y}_i}(y_i), \forall i$ equal to $\alpha$ such that $z_1 = b_1$ in order to satisfy $\mu_{\tilde{b}_1}(z_1) = \alpha$, where $b_1$ is defined in (2i). Since, all $\alpha$-cuts form a nested structure with respect to $\alpha$ as illustrated by Kaufmann (1975), then $\mu_{\tilde{x}_i}(x_i) \geq \alpha$ and $\mu_{\tilde{x}_i}(x_i) = \alpha$ have the same domain as $\mu_{\tilde{y}_i}(y_i) \geq \alpha$ and $\mu_{\tilde{y}_i}(y_i) = \alpha$. To find the membership function $\mu_{\tilde{b}_1}(z_1)$, it is sufficient to derive the lower and upper bounds of the $\alpha$-cuts of $\tilde{b}_1$. This is achieved by solving following crisp parametric nonlinear programming problems:

$$(\tilde{b}_1)_\alpha^L = \min \frac{\sum_{i=1}^n (x_i - \bar{x})(y_i - \bar{y})}{\sum_{i=1}^n (x_i - \bar{x})^2} \text{ subject to } (x_i)_\alpha^L \leq x_i \leq (x_i)_\alpha^U; (y_i)_\alpha^L \leq y_i \leq (y_i)_\alpha^U, \forall i \quad (7i)$$

$$(\tilde{b}_1)_\alpha^U = \max \frac{\sum_{i=1}^n (x_i - \bar{x})(y_i - \bar{y})}{\sum_{i=1}^n (x_i - \bar{x})^2} \text{ subject to } (x_i)_\alpha^L \leq x_i \leq (x_i)_\alpha^U; (y_i)_\alpha^L \leq y_i \leq (y_i)_\alpha^U, \forall i \quad (7ii)$$

where at least one $x_i$ and $y_i$ must hit the boundary of their $\alpha$-cuts to satisfy $\mu_{\tilde{b}_1}(z_1) = \alpha$. The objective functions of Equations (7i) and (7ii), $\bar{x} = \frac{\sum_{i=1}^n x_i}{n}$ and $\bar{y} = \frac{\sum_{i=1}^n y_i}{n}$ are crisp numbers rather than fuzzy numbers as all $x_i$ and $y_i$, $i = 1, \ldots, n$ are crisp numbers. Similarly, the membership function $\mu_{\tilde{b}_0}(z_0)$ can be obtained by deriving the lower and upper bounds of the $\alpha$-cut of $\tilde{b}_0$, which is solved by following parametric nonlinear programming problems:

$$(\tilde{b}_0)_\alpha^L = \min \frac{\sum_{i=1}^n y_i \sum_{i=1}^n x_i^2 - \sum_{i=1}^n x_i \sum_{i=1}^n x_i y_i}{n \sum_{i=1}^n x_i^2 - (\sum_{i=1}^n x_i)^2} \text{ subject to } (x_i)_\alpha^L \leq x_i \leq (x_i)_\alpha^U; (y_i)_\alpha^L \leq y_i \leq (y_i)_\alpha^U, \forall i \quad (8i)$$

$$(\tilde{b}_0)_\alpha^U = \max \frac{\sum_{i=1}^n y_i \sum_{i=1}^n x_i^2 - \sum_{i=1}^n x_i \sum_{i=1}^n x_i y_i}{n \sum_{i=1}^n x_i^2 - (\sum_{i=1}^n x_i)^2} \text{ subject to } (x_i)_\alpha^L \leq x_i \leq (x_i)_\alpha^U; (y_i)_\alpha^L \leq y_i \leq (y_i)_\alpha^U, \forall i \quad (8ii)$$

Each of the above two models is a nonlinear program with bound constraints, which can be solved effectively and efficiently by using nonlinear programming algorithms, such as sequential quadratic programming methods given by Bazarra, Sherali and Shetty (1993). The membership functions $\mu_{\tilde{b}_i}(z_i)$ are constructed as:

$$\mu_{\tilde{b}_i}(z_i) = \begin{cases} 0, (b_i)_{\alpha=1}^L \leq z_i, \\ L_i(z_i), (b_i)_{\alpha=0}^L \leq z_i \leq (b_i)_{\alpha=1}^L, \\ 1, (b_i)_{\alpha=1}^L \leq z_i \leq (b_i)_{\alpha=1}^U; i = 0,1 \\ R_i(z_i), (b_i)_{\alpha=1}^U \leq z_i \leq (b_i)_{\alpha=0}^U, \\ 0, (b_i)_{\alpha=0}^U \leq z_i \end{cases} \quad (9)$$

If the values of $(\tilde{b}_1)_\alpha^L$, $(\tilde{b}_1)_\alpha^U$, $(\tilde{b}_0)_\alpha^L$ and $(\tilde{b}_0)_\alpha^U$ cannot be solved analytically, then the numerical values for them at different level $\alpha$ can be collected to approximate the shapes of $L_i(z_i)$ and $R_i(z_i), i = 0,1$.

## 4. Fuzzy Regression Model with non-uniform spreads

Kao and Chyu (2002), (2003) gave the idea that the ideal regression coefficients in a fuzzy linear regression model should be crisp rather than fuzzy numbers, to avoid the increasing spread of estimated response. Further, a better fuzzy regression model can cope with the situation of decreasing or non-uniform spread. Here, we discuss the improved fuzzy linear regression model with better forecasting performances.

### 4.1 Crisp transformation

To find representative crisp values for $\tilde{b}_1$ and $\tilde{b}_0$, we defuzzify fuzzy numbers $\tilde{b}_1$ and $\tilde{b}_0$ to crisp values. Among the defuzzification approaches, center of area (COA) method is most commonly used technique used here for defuzzifying fuzzy regression coefficients to crisp ones as illustrated by Chen and Hwang (1992) and Ross (1997).

Let $(b_i)_c$ be defuzzified values of $\tilde{b}_i$. COA method calculates $(b_i)_c$ as:

$$(b_i)_c = \frac{\int_{(b_i)_{\alpha=0}^L}^{(b_i)_{\alpha=0}^U} z_i \mu_{\tilde{b}_i}(z_i) dz_i}{\int_{(b_i)_{\alpha=0}^L}^{(b_i)_{\alpha=0}^U} \mu_{\tilde{b}_i}(z_i) dz_i}, i = 0,1 \quad (10)$$

where, $\mu_{\tilde{b}_i}(z_i)$ is the membership functions of $\tilde{b}_i, i = 0,1$, respectively which can be developed as discussed previously. If analytical form of $\mu_{\tilde{b}_i}(z_i)$ cannot be obtained, then numerical methods of approximation such as trapezoidal rule can be used.

### 4.2 Non-uniform Error terms

From Equation (3) we have the error term, $\tilde{\varepsilon}_i = \tilde{y}_i - (\beta_0 + \beta_1 \tilde{x}_i), i = 1,\ldots,n$ (11). Now, $\beta_0$ and $\beta_1$ can be estimated by $(b_i)_c, i = 0,1$. Substituting $(b_i)_c$ into Equation (11), we have $\tilde{\varepsilon}_i = \tilde{y}_i - ((b_0)_c + (b_1)_c \tilde{x}_i), i = 1,\ldots,n$ (12). When $\tilde{x}_i$ and $\tilde{y}_i$ are fuzzy numbers, the error terms $\tilde{\varepsilon}_i$ are also fuzzy numbers. Assuming $\tilde{x}_i$ and $\tilde{y}_i$ as trapezoidal fuzzy numbers $(x_i^L, x_i^{M_1}, x_i^{M_2}, x_i^R)$ and $(y_i^L, y_i^{M_1}, y_i^{M_2}, y_i^R)$ respectively, $\tilde{\varepsilon}_i$ are also trapezoidal in nature. An estimate $\tilde{E} = (-l,0,0,r)$ for $\tilde{\varepsilon}_i, i = 1,\ldots,n$ leads to fixing the resulting spread of their estimated response such that $\tilde{E}_i = (-l_i,0,0,r_i)$. Consequently, fuzzy linear regression model with non-uniform spreads (NUS) is $(\tilde{y}_i)_{NUS} = (b_0)_c + (b_1)_c \tilde{x}_i + \tilde{E}_i$ $= (b_0)_c + (b_1)_c \tilde{x}_i + (-l_i,0,0,r_i); i = 1,\ldots,n$ (13). To minimize total error in estimation the best values of $l$ and $r$ should be found. Chen (2004) defined the error in estimation as difference between observed and estimated responses: $D_i = \int_{S(\tilde{y}_i) \cup S(\hat{\tilde{y}}_i)} |\mu_{\tilde{y}_i}(y) - \mu_{\hat{\tilde{y}}_i}(y)| dy, i = 1,\ldots,n$ (14) where, $S(\tilde{y}_i)$ and $S(\hat{\tilde{y}}_i)$ are supports of $\mu_{\tilde{y}_i}(y)$ and $\mu_{\hat{\tilde{y}}_i}(y)$ respectively. Taking Equation (14) as measure, a mathematical program is used to find the optimal values of spreads of each estimated response $l_i$ and $r_i$ such that total error in estimation is minimized. According to Equation (13), estimated response $\hat{\tilde{y}}_i$ can be represented as,

$$\hat{\tilde{y}}_i = (\hat{y}_i^L, \hat{y}_i^{M_1}, \hat{y}_i^{M_2}, \hat{y}_i^R)$$
$$= ((b_0)_c + (b_1)_c x_i^L + l_i, (b_0)_c + (b_1)_c x_i^{M_1}, (b_0)_c + (b_1)_c x_i^{M_2}, (b_0)_c + (b_1)_c x_i^R + r_i) \quad (15)$$

To cope with the situation of decreasing non-uniform spread for observed responses; here we assume the spread of each estimated response equal to that of its associated observed response, such that following equalities hold: $\hat{y}_i^R - \hat{y}_i^L = y_i^R - y_i^L, i = 1,2,\ldots,n$ (16). When $x_i$ is crisp, Equation (16) becomes $\hat{y}_i^R - \hat{y}_i^L = l_i + r_i = y_i^R - y_i^L, i = 1,2,\ldots,n$. The above constraints only limit value of the sum of left and right spreads for each estimated response. It is possible that Equation (16) is valid, but $l_i$ and $r_i$ is much smaller than it's observed left or right spreads respectively. Thus, for individual left or right spread of the estimated response, a lower bound should be considered [6]. $\hat{y}_i^{M_1} - \hat{y}_i^L \geq l_{\min}, i = 1,2,\ldots,n$ (17i); $\hat{y}_i^R - \hat{y}_i^{M_2} \geq r_{\min}, i = 1,2,\ldots,n$ (17ii) where, $l_{\min}$ and $r_{\min}$ are smallest left and right spreads of the observed responses respectively. Hence, incorporating Equations (16), (17i) and (17ii), the mathematical program to find the optimal values of $l_i^*$ and $r_i^*, i = 1,2,\ldots,n$ can be formulated as:

$$\min \sum_{i=1}^n D_i \text{ subject to } D_i = \int_{S(\tilde{y}_i) \cup S(\hat{\tilde{y}}_i)} |\mu_{\tilde{y}_i}(y) - \mu_{\hat{\tilde{y}}}(y)| \, dy, i = 1,\ldots,n;$$
$$\hat{\tilde{y}}_i = (b_0)_c + (b_1)_c \tilde{x}_i + \tilde{E}_i, i = 1,\ldots,n; \quad \tilde{E}_i = (-l_i, 0, 0, r_i), i = 1,\ldots,n;$$
$$\hat{y}_i^R - \hat{y}_i^L = y_i^R - y_i^L, i = 1,\ldots,n; \quad \hat{y}_i^{M_1} - \hat{y}_i^L \geq l_{\min}, i = 1,\ldots,n;$$
$$\hat{y}_i^R - \hat{y}_i^{M_2} \geq r_{\min}, i = 1,\ldots,n$$

**4.3 Fuzzy Inference for Forecasting**

Substituting $l_i^*$ and $r_i^*$ for $l_i$ and $r_i$ in Equation (13) respectively, the fuzzy regression model becomes $(\tilde{y}_i)_{NUS} = (b_0)_c + (b_1)_c \tilde{x}_i + (-l_i^*, 0, 0, r_i^*), i = 1,\ldots,n$ (19). This can be adopted to derive the estimated response for forecasting the associated observed response for specific and collected values of independent variables $\tilde{x}_i, i = 1,\ldots,n$. For predicting the response corresponding to the values other than these specific ones, Zimmermann (2001) applied the fuzzy inference system to fuzzy regression model. According to Cheng and Lee (1999) the fuzzy inference system is an important function approximation technique and has been applied to fields such as Expert Systems and Automatic Control. It consists of three components viz., rule base, database and reasoning mechanism. The rule base contains the selection of fuzzy if-then rules activated by certain value of interest, the database defines the membership functions adopted in the fuzzy if-then rules, and the inference procedure is called the fuzzy reasoning based on information aggregation from the activated fuzzy rules. Different types of fuzzy rules and aggregation resulting in several fuzzy inference systems have been used of which Mamdani fuzzy model and Tagaki-Sugeno fuzzy model are most important. In this work, Tagaki-Sugeno model with one input – one output is adopted for deriving the spread of predicted error terms. The output from a Tagaki-Sugeno fuzzy model is developed using a training data set or optimized by reducing training error.

Suppose that observed responses activated by independent variable of interest are $\tilde{y}_i^a, 1 = 1,\ldots,p$ and the associated error terms are $\tilde{E}_i^a = (-l_i^a, 0, 0, r_i^a), 1 = 1,\ldots,p$. Let $y_{estimated}$ denote the estimated response and

$e_{estimated}$ denote its estimated error term. Then each row of the membership function constitutes an if-then rule given as, $R_i$: if $y_{estimated} = \tilde{y}_i^a$, then $e_{estimated} = \tilde{E}_i^a$ $i = 1,\ldots,p$ (20). It is to be noted that fuzzy set of predicted error term, $\tilde{E}_{predicted}$ obtained from above may have irregular shapes. For all observations that are trapezoidal fuzzy numbers, it is preferable that the obtained estimated response is also a trapezoidal fuzzy number; which is achieved by transforming $\tilde{E}_{predicted}$ into a fuzzy number. Denoting the transformed predicted error term, $\tilde{E}_{estimated} = (l_{estimated}, m^1_{estimated}, m^2_{estimated}, r_{estimated})$, where $l_{estimated}$ and $r_{estimated}$ are taken as minimum and maximum of the possible $\tilde{E}_{predicted}$ values; $m^1_{estimated}$ and $m^2_{estimated}$ are calculated by using COA method. Thus, estimated response for $\tilde{x}_{estimated}$ becomes,

$$\tilde{y}_{NUS}^{estimated} = (b_0)_c + (b_1)_c \tilde{x}_{estimated} + (l_{estimated}, c^1_{estimated}, c^2_{estimated}, r_{estimated}) \quad (21)$$

## 5. Numerical Example

Consider the data given in the Table below. The spreads of the observed responses are 1, 1, 1, 1 and 5. By applying the non-uniform spread method, the membership functions of fuzzy regression coefficients are:

$$\mu_{\tilde{b}_1}(z_1) = \begin{cases} 0, 1.6 \le z_1, \\ \dfrac{10z_1 - 16}{8}, 1.6 \le z_1 \le 2.4, \\ 1, z_1 = 2.4, \\ \dfrac{3.2 - 10z_1}{8}, 2.4 \le z_1 \le 3.2, \\ 0, 3.2 \le z_1 \end{cases} \qquad \mu_{\tilde{b}_0}(z_0) = \begin{cases} 0, 0.7 \le z_0, \\ \dfrac{z_0 + 1.4}{2.1}, -1.4 \le z_0 \le 0.7, \\ 1, z_0 = 0.7, \\ \dfrac{2.8 - z_0}{2.1}, 0.7 \le z_0 \le 2.8, \\ 0, 2.8 \le z_0 \end{cases}$$

The regression model is $\tilde{y}_{NUS} = 0.6 + 2.4x + \tilde{E}_i, i = 1,\ldots,5$, where, $\tilde{E}_1 = (-0.6, 0, 0, 0.6)$, $\tilde{E}_2 = (-0.6, 0, 0, 0.6)$, $\tilde{E}_3 = (-0.6, 0, 0, 0.6)$, $\tilde{E}_4 = (-0.6, 0, 0, 0.6)$ and $\tilde{E}_5 = (-0.6, 0, 0, 0.6)$.

By applying the Kao and Chyu method, the regression model is $\tilde{y}_{KC} = 0.6 + 2.4x + (-1.2, 0, 0, 0.796)$. To compare the explanatory power Equation (14) is used. The Table 3 shows the errors in estimation for two methods. It is expected that for observations with dramatically non-uniform spread, the non-uniform model performs better than those with uniform spread.

| Observation Number | $[x_i, \bar{y}_i]$ | Errors in Estimation | |
|---|---|---|---|
| | | Two-stage | Non-uniform spread |
| 1. | [1, (2.0, 2.5, 2.5, 3.0)] | 0.456 | 0.356 |
| 2. | [2, (5.0, 5.5, 5.5, 6.0)] | 1.093 | 0.836 |
| 3. | [3, (6.0, 6.5, 6.5, 7.0)] | 0.789 | 0.836 |
| 4. | [4, (9.0, 9.5, 9.5, 10.0)] | 0.557 | 0.356 |
| 5. | [5, (9.0, 11.5, 11.5, 14.0)] | 1.586 | 0.000 |
| Total Error | | 4.480 | 2.384 |

**Table 1: The numerical data and estimation errors**

## 6. Conclusion

Regression analysis provides a great deal of information into forecasting problems in Operations Research applications. Some of previous works on fuzzy regression coefficients of increasing spreads for the estimated fuzzy responses as the independent variable increase in magnitude, and obtain crisp regression coefficients with uniform spreads which are not suitable for general cases. Although some obtain crisp regression coefficients and uniform spread, they cannot deal with the situation where the spreads of the observed responses are actually non-uniform. Here, the regression coefficients are calculated as crisp values and the spreads of fuzzy error terms are non-uniform. The models proposed earlier with an increasing spread or constant spread are suitable for cases when observed responses have increasing spread or constant spread. The non-uniform spread fuzzy linear regression model resolves the problem of wider spreads of estimated response for larger values of independent variables in fuzzy regression analysis and deals with non-uniform spreads effectively. The method is based on the extension principle which also provides the membership function of least-squares estimate of the regression coefficient, thereby conserving the fuzziness of observations. Further, the method has greater explanatory power and forecasting accuracy. Finally, the method for constructing the membership function of fuzzy regression coefficient completely conserves the fuzziness of input information. The numerical example illustrates the strength of non-uniform method in terms of better explanatory power than earlier studies. The non-uniform method can also be applied to multiple fuzzy regression problems.